\documentclass[11pt]{article}

\usepackage[preprint]{acl}

\usepackage{times}
\usepackage{latexsym}
\usepackage{amsmath}
\usepackage{booktabs}
\usepackage[table]{xcolor}
\usepackage{multirow}
\usepackage{placeins}
\usepackage{wrapfig}

\usepackage[T1]{fontenc}

\usepackage[utf8]{inputenc}

\usepackage{microtype}

\usepackage{inconsolata}

\usepackage{geometry}
\usepackage{tcolorbox}
\tcbuselibrary{skins, breakable}

\usepackage{graphicx}

\usepackage{tabularx}
\usepackage{makecell}
\usepackage{booktabs}
\usepackage{ragged2e}
\usepackage{amssymb}
\usepackage{array}
\usepackage{enumitem}
\definecolor{headergray}{gray}{0.92} 

\newcolumntype{L}{>{\RaggedRight\arraybackslash}X}
\title{RealMem: Benchmarking LLMs in Real-World Memory-Driven Interaction}

\author{
  \textbf{Haonan Bian\textsuperscript{1,*}},
  \textbf{Zhiyuan Yao\textsuperscript{2,*}},
  \textbf{Sen Hu\textsuperscript{3,*,\textdagger}},
   \textbf{Zishan Xu\textsuperscript{4}},
  \textbf{Shaolei Zhang\textsuperscript{5}},
\\
 \textbf{Yifu Guo\textsuperscript{6}},
  \textbf{Ziliang Yang\textsuperscript{1}},
  \textbf{Xueran Han\textsuperscript{5}},
  \textbf{Huacan Wang\textsuperscript{7}},
  \textbf{Ronghao Chen\textsuperscript{3}}
\\
\\
  \textsuperscript{1}Xidian University,
  \textsuperscript{2}Zhejiang University,
  \textsuperscript{3}Peking University,
  \textsuperscript{4}Shanghai Jiao Tong University,
\\
  \textsuperscript{5}Renmin University of China,
  \textsuperscript{6}Sun Yat-sen University,
  \textsuperscript{7}University of the Chinese Academy of Sciences
\\
  \small{
    \textsuperscript{*}Equal contribution.
    \textsuperscript{\textdagger}Corresponding author: \href{mailto:husen@pku.edu.cn}{husen@pku.edu.cn}
  }
}

\begin{document}

\maketitle

\begin{abstract}
As Large Language Models (LLMs) evolve from static dialogue interfaces to autonomous general agents, effective memory is paramount to ensuring long-term consistency. However, existing benchmarks primarily focus on casual conversation or task-oriented dialogue, failing to capture ``long-term project-oriented'' interactions where agents must track evolving goals. To bridge this gap, we introduce \textbf{RealMem}, the first benchmark grounded in realistic project scenarios. RealMem comprises over 2,000 cross-session dialogues across eleven scenarios, utilizing natural user queries for evaluation. We propose a synthesis pipeline that integrates Project Foundation Construction, Multi-Agent Dialogue Generation, and Memory and Schedule Management to simulate the dynamic evolution of memory. Experiments reveal that current memory systems face significant challenges in managing the long-term project states
 and dynamic context dependencies inherent in real-world projects.
Our code and datasets are available at \url{https://github.com/AvatarMemory/RealMemBench}.
\end{abstract}
\section{Introduction}

In recent years, large language models (LLMs) have achieved remarkable progress across multiple dimensions~\cite{du2025graphoracle,du2025graphmaster,guo2025octopusagenticmultimodalreasoning,luo2025codetestcasesenough}. There is a growing consensus that AI agents must evolve from turn-based chatbots into long-term companions capable of sustained, context-aware collaboration~\cite{tran2025multi,sapkota2025ai}.
In this paradigm shift, memory systems have emerged as a foundational component~\cite{hu2025memory}. 

Robust memory is essential not only for practical applications such as  personalized chatbots~\cite{li2025hello} and financial analysis~\cite{zhang2024multimodal}, but also as a key enabler toward Artificial General Intelligence (AGI)~\cite{fang2025comprehensive}.

\begin{figure}[t]
    \centering
    \includegraphics[width=\linewidth]{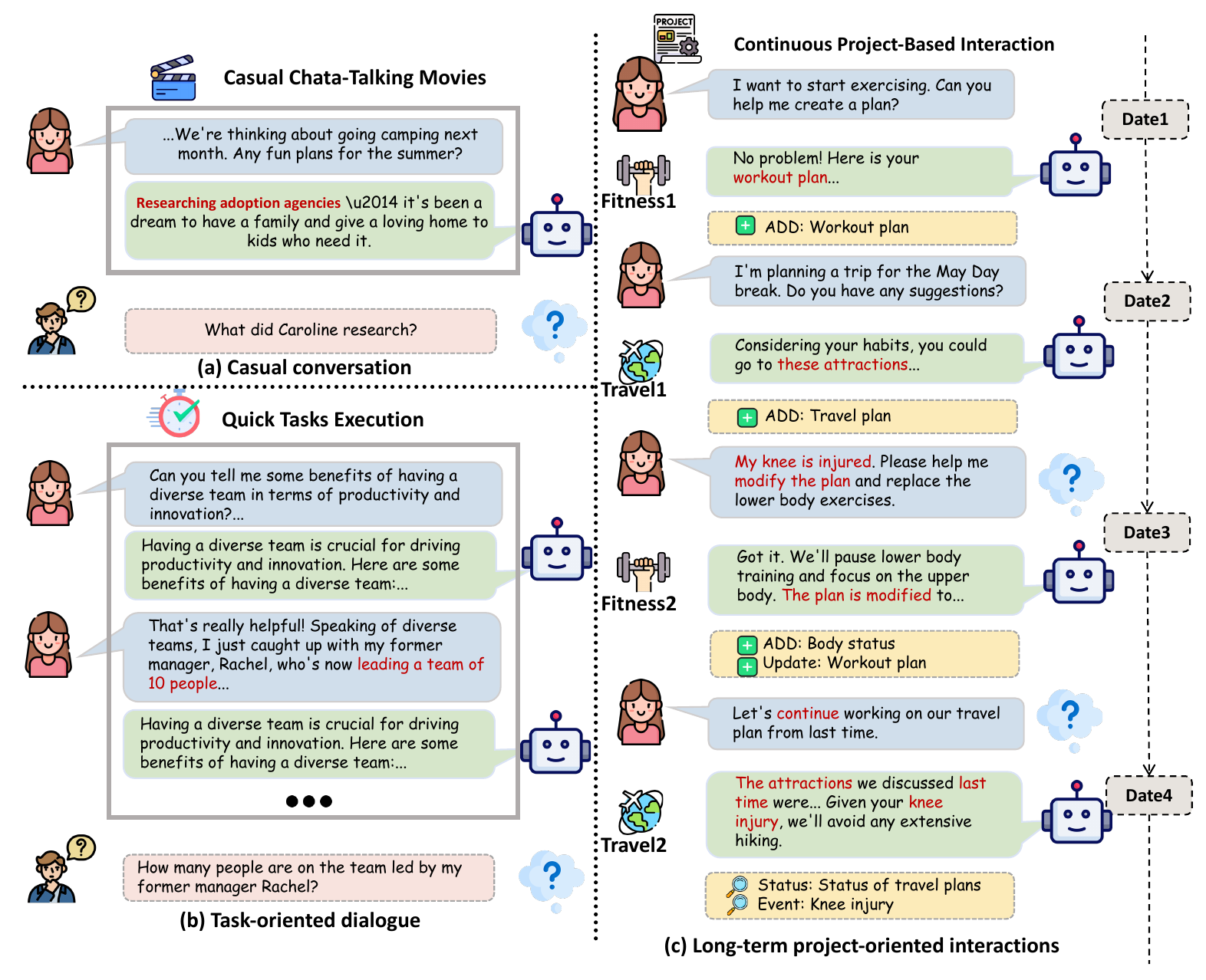}
    \caption{Comparison of three interaction paradigms in human--agent interactions: 
    (a) casual conversation, 
    (b) task-oriented dialogue, and 
    (c) long-term project-oriented interactions spanning multiple sessions with interleaved projects and evolving context.}
    \label{fig:compare_diglogue_type}
\end{figure}

\begin{table*}[t]
\centering
\small
\renewcommand{\arraystretch}{1.35} 

\renewcommand{\tabularxcolumn}[1]{m{#1}}

\newcolumntype{C}{>{\centering\arraybackslash}X}

\begin{tabularx}{\linewidth}{l C C C C}
\toprule
\textbf{Benchmark} & \textbf{RealMem} & \textbf{HaluMem} & \textbf{LongMemEval} & \textbf{LoCoMo} \\
\midrule
\textbf{Dialogue Type} & 
Long-term project-oriented & 
Persona Creation & 
Task-oriented & 
Casual conversation \\
\midrule
\textbf{QA Timing} & 
Interleaved within sessions & 
After each session & 
After all sessions & 
After all sessions \\
\midrule
\textbf{Query Source} & 
Natural user queries & 
External queries & 
External queries & 
External queries \\
\midrule
\textbf{Proactive Alignment} & $\checkmark$ & $\times$ & $\times$ & $\times$ \\
\midrule
\textbf{Project States Memory} & $\checkmark$ & $\times$ & $\times$ & $\times$ \\
\midrule
\textbf{Max Memory Items} & 20 & 9 & 6 & 19 \\
\midrule
\textbf{Memory Content} & 
\makecell{Project States, \\ Schedules, \\ Personas} & 
\makecell{Personas, \\ Isolated Events, \\ Relationships} & 
\makecell{Personas, \\ Isolated Events} & 
\makecell{Personas, \\ Isolated Events} \\
\bottomrule
\end{tabularx}
\caption{Comparison of representative memory benchmarks. RealMem distinguishes itself by introducing \textbf{proactive alignment} and \textbf{project state memory} into long-term interactions, shifting the evaluation paradigm from isolated fact-checking to continuous, \textbf{natural dialogue} progression.}
\label{tab:memory_benchmarks_lines}
\end{table*}

Real-world memory-driven interaction empowers agents to maintain coherent, long-term collaboration~\cite{zhang2025survey}.
We posit that long-term project-oriented interactions is governed by four fundamental imperatives: (1) Endogenous Query Nature: Queries arise organically from task progression, rather than appearing as isolated fact-checking questions. (2) Interleaved Distribution: User inquiries are interwoven across fragmented sessions.
(3) Dynamic State Evolution: The interaction environment is inherently non-stationary, demanding continuous synchronization of memory with constantly evolving information states~\cite{majumder2023clin,zhang2025agentic}. (4) Proactive Contextual Alignment: Agents should proactively resolve ambiguous intents by leveraging memory details (e.g., schedule management), and simultaneously maintain granular state updates that capture situational transitions rather than simplistic factual overwrites.

To contextualize these requirements, we categorize user--agent interactions into three paradigms, as illustrated in Figure~\ref{fig:compare_diglogue_type}: \emph{casual conversation}, \emph{task-oriented dialogue}, and \emph{long-term project-oriented interactions}. 
We argue that the third paradigm constitutes a crucial component of real-world memory-driven interactions.
For example, in a fitness program spanning six months, an AI assistant can function as a personal trainer, providing continuous guidance by leveraging accumulated memories from past interactions.
Existing dialogue memory benchmarks primarily focus on the first two interaction paradigms.
For example, LoCoMo~\cite{maharana2024evaluating} is limited to simulating human--human social chit-chat, while LongMemEval~\cite{wu2024longmemeval} approximates task dialogues via artificial ``needle-in-a-haystack'' tests. Consequently, their information flow remains discrete and episodic, failing to reflect the coherence of real-world workflows.

To bridge this gap, we introduce \textbf{RealMem}, a benchmark explicitly constructed to evaluate Real-World Memory-Driven Interaction through the lens of long-term projects. Grounded in eleven representative scenarios requiring robust long-term memory, RealMem comprises over 2,000 cross-session dialogues, thereby shifting the evaluation paradigm from isolated fact retrieval to project-centric memory utilization. RealMem evaluates agents using natural user queries that are organically interwoven across fragmented sessions. Our core focus is to assess its capacity to leverage accumulated memory and maintain the coherent thread of a project to fulfill user requests within a realistic and evolving context.

\begin{figure*}[t]
    \centering
    \includegraphics[width=0.95\linewidth]{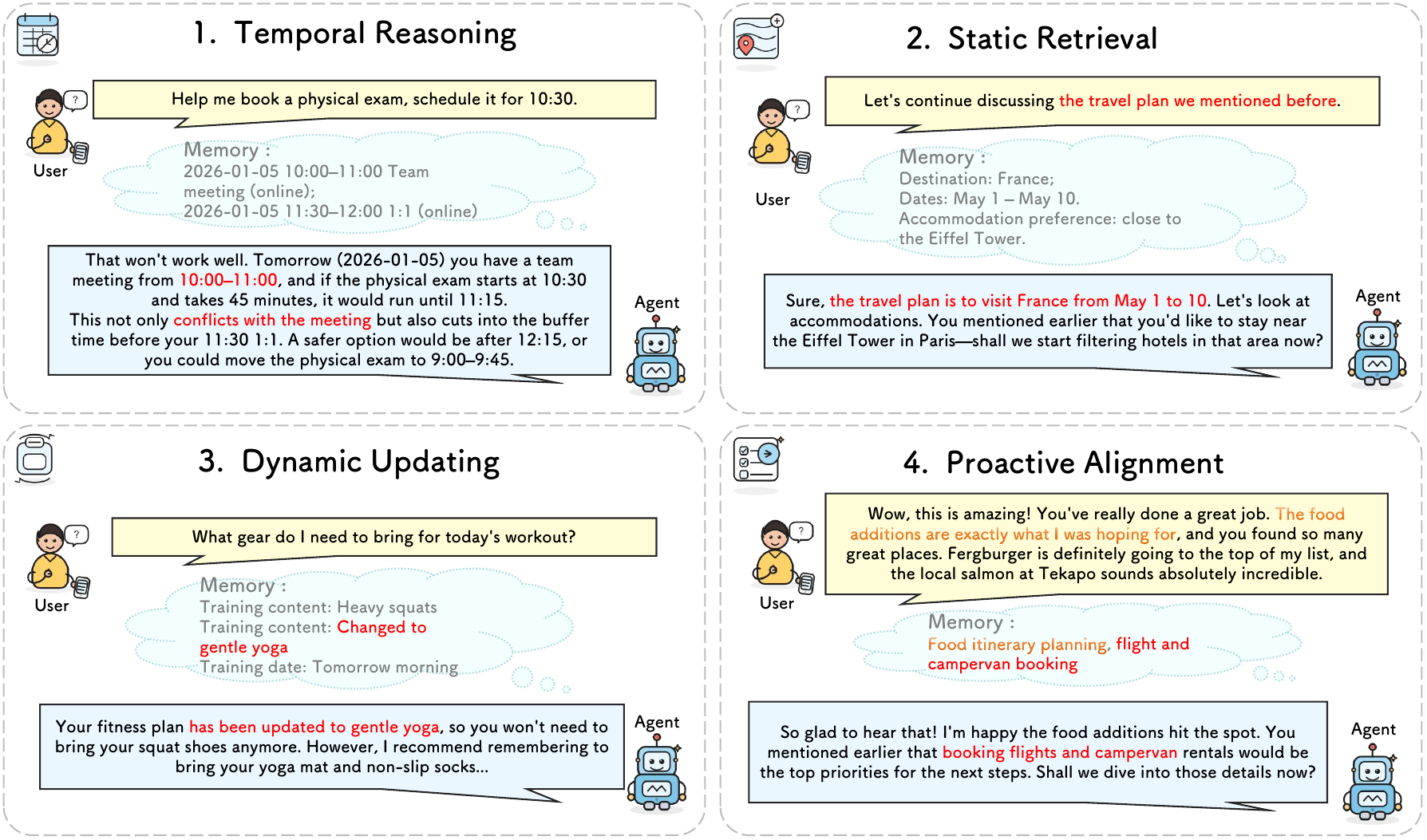}
    \caption{Examples of four query types in \textsc{RealMem}: (1) \textbf{Temporal Reasoning} resolves temporal constraints and schedule conflicts; (2) \textbf{Static Retrieval} ensures continuity by recalling accumulated context; (3) \textbf{Dynamic Updating} synchronizes memory with evolving project states; and (4) \textbf{Proactive Alignment} leverages user memory to anticipate implicit intents and goals.}
    \label{fig:question-type}
\end{figure*}

We design a three-stage synthesis pipeline comprising \emph{Project Foundation Construction}, \emph{Multi-Agent Dialogue Generation}, and \emph{Memory and Schedule Management}. This pipeline simulates the continuous evolution of long-term interactions, ensuring that memory is not predefined but dynamically emerges and evolves alongside the dialogue trajectory. Furthermore, we define a diverse set of query types (as illustrated in Figure~\ref{fig:question-type}) to rigorously evaluate the agent's memory system. 
Through extensive evaluation on RealMem, we demonstrate that existing agent memory systems are inadequate for the dynamic requirements of long-term project-oriented interactions.

Our contributions are summarized as follows: 
\begin{itemize} 

\item We identify a critical gap between existing memory benchmarks and real-world memory-driven interaction: long-term project-oriented scenarios. To address this gap, we introduce \textbf{RealMem}, which shifts the evaluation focus from post hoc fact retrieval to the proactive use of memory within ongoing conversations to generate more effective responses.

\item We propose a three-stage synthesis pipeline for constructing long-term project-oriented dialogues. This framework ensures global logical coherence across sessions and projects, while supporting the dynamic evolution of fine-grained memories.

\item Extensive evaluations reveal that state-of-the-art agent memory systems struggle to maintain \textbf{coherent project threads}. Our findings expose a critical gap in existing models' ability to proactively align with evolving contexts in long-term interactions.
\end{itemize}

\section{Related Work}

\subsection{Long-Term Memory Benchmarks}
A number of benchmarks~\cite{wei2025evo, jiang2025personamem} have been proposed to evaluate the memory capabilities of LLM-based agents. In this work, we focus on representative benchmarks most relevant to dialogue-based settings, as summarized in Table~\ref{tab:memory_benchmarks_lines}.

LoCoMo~\cite{maharana2024evaluating} established an end-to-end evaluation paradigm for memory retention via question answering over ultra-long contexts, though it primarily assesses static information recall without explicit updates. LongMemEval~\cite{wu2024longmemeval} extended this framework by incorporating memory updating mechanisms to rigorously quantify retrieval precision and knowledge consistency across multi-year spans. HaluMem~\cite{chen2025halumem} utilizes injected persona profiles to assess hallucinations over extended timeframes, specifically measuring the consistency of generated responses against pre-defined facts.
However, these methods predominantly rely on externally constructed QA tasks to evaluate isolated factual details, often overlooking the dynamic state consistency essential for practical applications.

In contrast, RealMem targets long-term project-oriented interactions. It requires agents not only to recall static facts but to actively track evolving project states and handle interleaved queries that arise naturally across fragmented sessions.

\subsection{Agent Memory Systems}

To support effective long-term interaction, existing approaches commonly adopt RAG-based mechanisms for memory retrieval~\cite{gutierrez2025rag,Thang2025marag}, and a variety of agent memory systems have been proposed. These systems typically distinguish between experiential memory~\cite{tang2025agent,ouyang2025reasoningbank,wang2024recmind,kim2025principlessyntheticstrategymemory}, derived from reflections on conversational history, and factual memory~\cite{li2025memos,fang2025lightmem, wang2025mirixmultiagentmemoryllmbased, rasmussen2025zep}, which records knowledge acquired through agent--user and agent--environment interactions. A-mem~\cite{xu2025mem} proposes an agentic memory organization framework to improve memory effectiveness in long-term interactions. Mem0~\cite{chhikara2025mem0} introduces a scalable memory-centric architecture with graph-based representations to capture relational structures across dialogue elements. 
MemoryOS~\cite{kang2025memoryosaiagent} adopts operating-system-inspired memory management to improve retrieval and update efficiency. Graph Memory~\cite{hu2026doesmemoryneedgraphs} further enhances the knowledge graph’s capacity to represent memory by allowing entity descriptions to record relevant events and remain up to date.

\section{Method}

\begin{figure*}[t]
  \includegraphics[width=\linewidth]{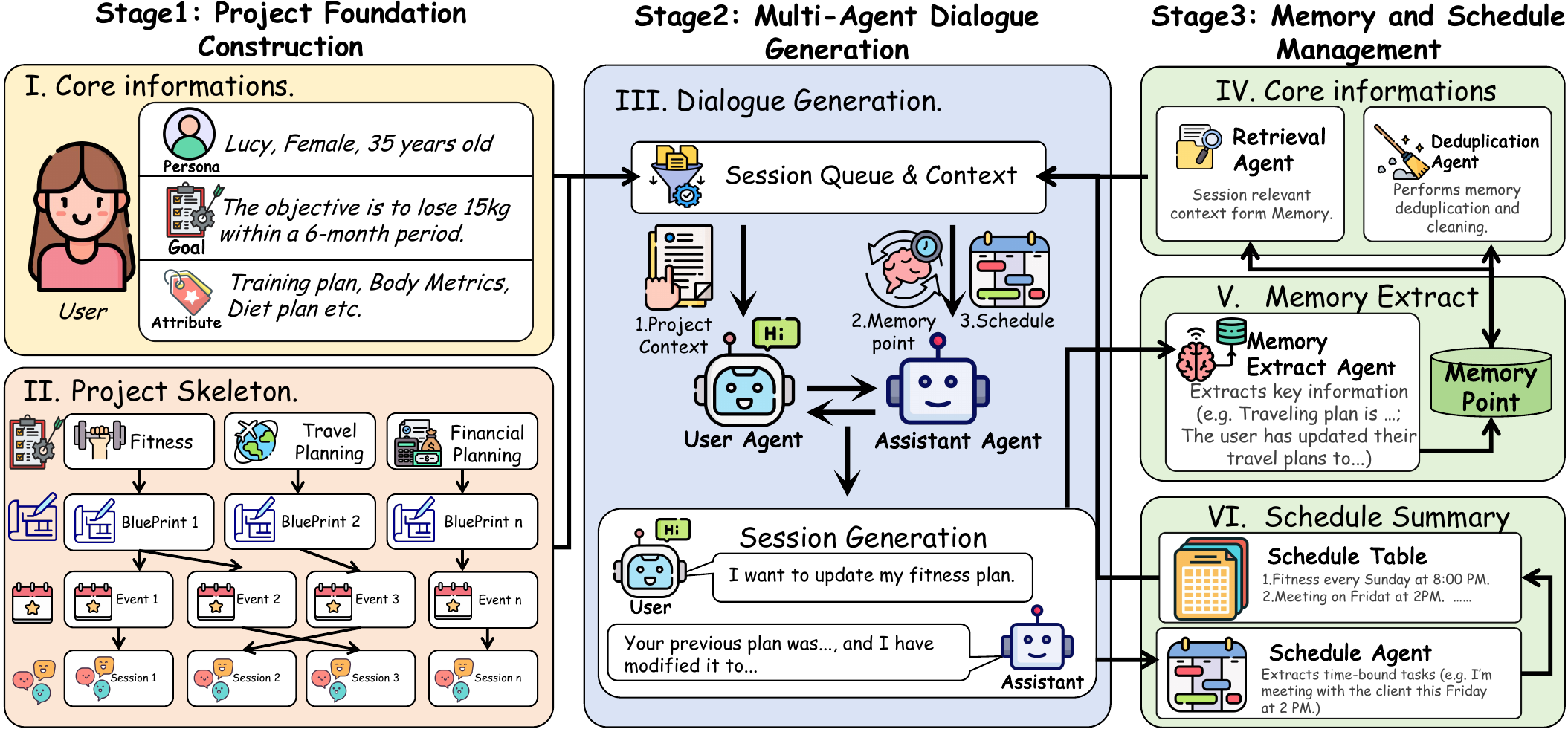}
  \caption{Overview of the data synthesis framework. The pipeline consists of three cascaded stages: (1) \textbf{Project Foundation Construction}, which initializes user personas and hierarchical project skeletons (i.e., blueprints, events, and sessions); (2) \textbf{Multi-Agent Dialogue Generation}, where the User Agent and Assistant Agent simulate interactions based on the session queue and dynamic context; and (3) \textbf{Memory and Schedule Management}, which iteratively retrieves, updates, and deduplicates memory points and schedule tables to ensure long-term consistency.}
  \label{fig:framework}
\end{figure*}

To evaluate memory systems in complex and evolving contexts, we introduce the \textbf{RealMem} benchmark. Inspired by recent advances in agentic frameworks (Figure~\ref{fig:framework}) that leverage multi-agent collaboration for complex problem solving~\cite{huang2024agentcoder,liu2024toolace,tang2025synthesizing}, RealMem simulates realistic user behavior in which multiple long-term projects are managed concurrently within a unified ``SuperApp'' environment. Its design is structured around three tightly coupled stages---static project foundation construction, dynamic interaction generation, and closed-loop memory feedback---which together enable systematic assessment of an agent’s ability to utilize, update, and maintain dynamic memory over extended interactions.

\subsection{Project Foundation Construction}

The \textbf{Project Foundation Construction} phase initializes the static context and structural scaffolding required for subsequent dialogue generation. As long-horizon generation demands explicit planning to preserve global coherence~\cite{xia2025storywriter}, we adopt a hierarchical strategy to construct a \textit{Project Skeleton}, following a ``blueprint-first'' paradigm that has proven effective in recent multi-turn data generation frameworks~\cite{prabhakar2025apigen,chen2025halumem}.

Specifically, this phase establishes a user \textit{Persona} to encode demographics for behavioral consistency, and a \textit{Project Goal} to specify a quantitative objective (e.g., ``losing 15\,kg in six months'') as a clear long-term target. Crucially, we introduce \textit{Project Attributes} as dynamic state variables that serve as the core mechanism for project modeling, tracking temporal evolution to capture changes in user progress and contextual state.

To reflect realistic user needs and assess the generality of our framework, we curate eleven representative scenarios spanning four application domains. For each scenario, we identify a set of core attributes—ranging from schedule adjustments to emotional regulation—to capture the multifaceted nature of long-term human--AI interaction.

The generation process proceeds in three stages. First, a \textit{Project Blueprint} is constructed to outline high-level milestones. Second, an \textit{Event List} is generated to encode causal dependencies among milestones. Third, \textit{Session Summaries} are produced to guide individual dialogue sessions. This hierarchical decomposition ensures that local interactions remain aligned with the global project narrative, mitigating fragmentation commonly observed in long-context generation~\cite{xia2025storywriter,prabhakar2025apigen}. Finally, to simulate concurrent task management, session summaries from multiple projects are aggregated and interleaved to form a unified session queue.

\subsection{Multi-Agent Dialogue Generation}

Inspired by recent advances in multi-agent data synthesis~\cite{liu2024toolace,xia2025storywriter}, we adopt a dual-agent framework consisting of a \textit{User Agent} and an \textit{Assistant Agent} to simulate realistic user--agent interactions. Compared to \emph{one-shot generation of multi-turn dialogues}, this simulation-based paradigm provides finer-grained control over interaction dynamics and policy constraints, as demonstrated in prior work such as IntellAgent~\cite{levi2025intellagent}.

To ensure long-term consistency, each dialogue session is generated under a structured context spanning three dimensions: (1) static project background, (2) memory points extracted from historical interactions, and (3) established schedules. Dialogue generation proceeds at the session level, with sessions sequentially drawn from an interleaved queue that simulates concurrent project management.

For the User Agent, the context includes information about the current event together with a summary of the immediately preceding event, enabling coherent progression awareness without exposing future plans. Importantly, we restrict the User Agent’s input to session-relevant summaries only, thereby enforcing explicit \textit{task boundaries} and preventing premature reasoning over future tasks. In contrast, the Assistant Agent is provided with all memory points relevant to the current session to support informed decision making. To further mitigate temporal inconsistencies frequently observed in LLM-based generation (e.g., scheduling conflicts across projects), we incorporate a \textit{Global Schedule} into the Assistant Agent’s context. This design substantially improves temporal alignment and logical coherence across long-term, multi-project interactions.

\subsection{Memory and Schedule Management}

This stage processes the generated dialogue data to form a closed feedback loop. To this end, our framework employs a set of specialized agents to validate, consolidate, and update the system state, ensuring that memory evolves accurately and consistently over time.

Specifically, a \textit{Memory Extraction Agent} parses raw dialogues to identify salient facts and converts them into structured memory points. In parallel, a \textit{Schedule Agent} detects tasks with explicit temporal attributes (e.g., scheduled meetings or deadlines) and updates the global schedule accordingly. To improve storage efficiency and reduce noise, we introduce a \textit{Deduplication Agent} that performs semantic-level deduplication over the memory base, removing redundant or overlapping entries.

In subsequent dialogue generation cycles, the \textit{Session Summary} is used as a query to retrieve relevant information from the optimized memory base, which is then provided as contextual input to the Assistant Agent. This closed-loop design enables continuous memory refinement and supports coherent long-term interactions across sessions.

\section{Evaluation}

To comprehensively assess retrieval quality, we adopt a hybrid evaluation strategy that combines standard quantitative metrics with LLM-based semantic judgments~\cite{liu2023g}.

For retrieval performance, we report \textbf{Recall@k} and \textbf{NDCG@k} ($k \in \{10, 20\}$) to measure strict ranking accuracy. To capture semantic relevance beyond exact lexical matching, we further introduce LLM-based metrics: \textbf{Mem Recall}, which assesses the semantic coverage of relevant information, and \textbf{Mem Helpful}, which evaluates the practical usefulness of the retrieved context for answering the query.

For generation quality, we employ the \textbf{QA Score}, which is derived from a response consistency rubric (ranging from conflicting to fully state-aligned). This metric explicitly evaluates whether the agent correctly incorporates the user's dynamic state, rather than merely producing fluent but context-agnostic responses.

\begin{table*}[!t]
\centering
\resizebox{\textwidth}{!}{%
\begin{tabular}{lcccccccc}
\toprule
\multirow{2}{*}{\textbf{Method}} & \multicolumn{4}{c}{\textbf{Value = Memory}} & \multicolumn{4}{c}{\textbf{Value = Session}} \\
\cmidrule(lr){2-5} \cmidrule(lr){6-9}
 & \textbf{QA-4o-Mini} & \textbf{QA-4o} & \textbf{Mem Rec.} & \textbf{Mem Help.} & \textbf{QA-4o-Mini} & \textbf{QA-4o} & \textbf{Mem Rec.} & \textbf{Mem Help.} \\
\midrule
Graph Mem & 0.474 & 0.497 & 0.490 & 0.598 & \textbf{0.539} & 0.567 & \textbf{0.608} & \textbf{0.708} \\
Mem0 & 0.449 & 0.514 & 0.529 & \textbf{0.634} & 0.526 & \textbf{0.609} & 0.594 & 0.666 \\
A-mem & 0.416 & 0.492 & 0.455 & 0.513 & 0.504 & 0.606 & 0.590 & 0.655 \\
MemoryOS & \textbf{0.490} & \textbf{0.567} & \textbf{0.532} & 0.606 & -- & -- & -- & -- \\
\midrule
\color{gray}Oracle & \color{gray}0.683 & \color{gray}0.804 & \color{gray}0.993 & \color{gray}0.922 & \color{gray}0.580 & \color{gray}0.696 & \color{gray}0.767 & \color{gray}0.845 \\
\bottomrule
\end{tabular}%
}
\caption{Generation performance comparison. We report \textbf{QA Score} (on \textbf{GPT-4o-mini} and \textbf{GPT-4o}), \textbf{Mem Recall}, and \textbf{Mem Helpful} scores. ``Memory'' denotes memory-only context, while ``Session'' includes full session context. \textbf{Bold} indicates the best performance among non-oracle methods. Note that Value = session metrics are not applicable to MemoryOS due to its inherent mechanism.}
\label{tab:generation_results}
\end{table*}
\section{Experiment}
\subsection{Experiment Setup}

\begin{table}[htbp]
\centering
\begin{tabular}{lcccc}
\toprule
\textbf{Method} & \textbf{R@10} & \textbf{R@20} & \textbf{NDCG@10} & \textbf{NDCG@20} \\
\midrule
Graph Mem & \textbf{0.6344} & 0.7017 & \textbf{0.5654} & \textbf{0.5826} \\
Mem0      & 0.6055 & 0.6515 & 0.5251 & 0.5398 \\
A-mem     & 0.5760 & \textbf{0.7235} & 0.5211 & 0.5560 \\
\bottomrule
\end{tabular}
\caption{Retrieval performance evaluated on session-level context. Metrics include Recall@K (R@K) and NDCG@K. \textbf{Bold} indicates the best performance.}
\label{tab:retrieval_results}
\end{table}

We conduct a systematic evaluation of representative state-of-the-art memory systems
on our proposed RealMem benchmark, including Mem0~\cite{chhikara2025mem0}, A-mem~\cite{xu2025mem}, MemoryOS~\cite{kang2025memoryosaiagent}, and Graph Memory~\cite{hu2026doesmemoryneedgraphs}.
For A-mem, MemoryOS, and Graph Memory, we strictly follow the embedding models
and hyperparameter settings recommended in their original papers; for Mem0,
we use text-embedding-3-small~\cite{openai_text_embedding_3_small_docs} as its default embedding model.
We utilize GPT-4o-mini for memory extraction. For the answer generation phase, we employ both GPT-4o-mini and GPT-4o~\cite{hurst2024gpt} to assess performance across different backbones. Finally, to ensure consistency, we adopt GPT-4o as the LLM-as-a-Judge for all evaluations.

During the answer generation stage, we evaluate each method under two
independent context construction settings.
In the memory-only setting ( Value = memory ), the model is provided solely with the
Top-20 memory entries retrieved by the memory system.
In the session-based setting ( Value = session ), memory entries are not directly
used; instead, the model receives the corresponding Top-5 original dialogue
sessions associated with the retrieval results, in order to assess generation
performance based on session-level context.

We note that MemoryOS internally maintains memory at the page
level and does not support alignment with original dialogue sessions; as a result, it cannot be evaluated under the session-based setting.

\subsection{Overall Evaluation}

The results on generation tasks offer critical insights into the architectural requirements for long-term project agents.
First, under the \textbf{Memory-only} setting, the superior performance of \textbf{MemoryOS} validates the effectiveness of hierarchical memory architectures (e.g., STM, MTM, and LTM). It demonstrates that effective compression and indexing of key information allow agents to maintain high accuracy even without full session history. 
Second, when \textbf{Session} context is available, \textbf{Graph Memory} achieves the highest scores. This indicates that the \textbf{RealMem} benchmark involves intricate entity relationships, where graph structures excel at capturing complex dependencies, thereby providing more precise context than linear history alone. 

\begin{table*}[t]
\centering
\small
\setlength{\tabcolsep}{3.5pt} 
\renewcommand{\arraystretch}{1.2}

\begin{tabular}{l cc cc cc cc | cc cc}
\toprule
\multirow{3}{*}{\textbf{Method}} 
& \multicolumn{8}{c|}{\textbf{Memory Category}} 
& \multicolumn{4}{c}{\textbf{Session Type}} \\
\cmidrule(lr){2-9} \cmidrule(lr){10-13}
& \multicolumn{2}{c}{Dyn. Inc.} 
& \multicolumn{2}{c}{Dyn. Upd.} 
& \multicolumn{2}{c}{Proac. Align.} 
& \multicolumn{2}{c|}{Temp. Reas.} 
& \multicolumn{2}{c}{Single Sess.} 
& \multicolumn{2}{c}{Multi Sess.} \\
\cmidrule(lr){2-3}\cmidrule(lr){4-5}\cmidrule(lr){6-7}\cmidrule(lr){8-9} \cmidrule(lr){10-11}\cmidrule(lr){12-13}
& QA & Rec. & QA & Rec. & QA & Rec. & QA & Rec. & QA & Rec. & QA & Rec. \\
\midrule
A-mem        
& 0.404 & 0.465 & 0.453 & 0.427 & 0.483 & 0.436 & 0.278 & 0.361 
& 0.421 & 0.454 & 0.409 & 0.457 \\

Mem0         
& 0.436 & 0.529 & 0.462 & 0.553 & 0.538 & \textbf{0.506} & 0.347 & 0.539 
& 0.457 & 0.542 & 0.437 & \textbf{0.509} \\

MemoryOS     
& \textbf{0.478} & \textbf{0.536} & \textbf{0.521} & \textbf{0.557} & \textbf{0.571} & 0.490 & 0.319 & 0.448 
& \textbf{0.505} & \textbf{0.565} & \textbf{0.468} & 0.482 \\

Graph Mem 
& 0.470 & 0.500 & 0.483 & 0.512 & 0.513 & 0.392 & \textbf{0.375} & \textbf{0.559} 
& 0.483 & 0.516 & 0.461 & 0.451 \\
\bottomrule
\end{tabular}

\caption{Overall performance comparison (QA Score and Recall). Methods are listed in rows, while memory categories and session types are arranged in columns. Abbreviations used in the header: \textbf{Dyn. Inc.} (Dynamic Incremental), \textbf{Dyn. Upd.} (Dynamic Updating), \textbf{Proac. Align.} (Proactive Alignment), and \textbf{Temp. Reas.} (Temporal Reasoning).}
\label{tab:combined-results}
\end{table*}

Furthermore, a cross-model comparison reveals the challenging nature of realistic project scenarios. The QA accuracy of \textbf{GPT-4o} is consistently higher than that of \textbf{GPT-4o-mini}. Crucially, even with Oracle memory, GPT-4o-mini exhibits limited performance. This suggests that solving the complex queries in RealMem necessitates not only accurate retrieval but also high-performance foundation models with strong reasoning capabilities.
Finally, the substantial gap between all methods and the \textbf{Oracle} (e.g., Oracle Recall reaches 0.993) highlights significant room for improvement in capturing long-term context dependencies. This underscores a key finding: memory can effectively replace raw session history only when retrieval quality is sufficiently close to the theoretical upper bound.

The retrieval results further clarify the correlation between retrieval metrics and downstream generation quality in the RealMem benchmark. 
While \textbf{A-mem} achieves the highest Recall@20 (0.7235), its lower NDCG scores suggest it retrieves a broader but noisier set of information. Conversely, \textbf{Graph Memory} dominates in NDCG metrics (0.5654 @10), aligning with its superior generation performance in Table~\ref{tab:retrieval_results}. 
This discrepancy highlights a critical characteristic of the benchmark: for long-term project dialogues, \textbf{precision and ranking quality} (NDCG) are more decisive than mere coverage (Recall). High recall with low precision introduces noise that distracts the LLM, whereas high NDCG ensures that the most relevant context is prioritized, directly translating to better response quality.

\subsection{Performance on Different Question Types}

Table~\ref{tab:combined-results} presents a fine-grained performance analysis across different memory categories and session types. We summarize the key observations as follows:

Our results highlight distinct architectural strengths across different complexity dimensions. \textbf{MemoryOS} demonstrates robust superiority in handling dynamic information, achieving the highest scores in both \textit{Static Retrieval} and \textit{Dynamic Updating} categories (e.g., $0.521$ QA). This validates the effectiveness of its hierarchical architecture in tracking evolving user states and managing fragmented updates. Conversely, \textbf{Graph Memory} dominates the \textit{Temporal Reasoning} category (QA $0.375$, Recall $0.559$).

Regarding session types, performance across all models generally declines in \textit{Multi Session} settings compared to \textit{Single Session}, highlighting the inherent difficulty of maintaining long-range context. Nevertheless, \textbf{MemoryOS} maintains leading QA performance in both settings (Single: $0.505$, Multi: $0.468$), demonstrating its robustness in handling diverse interaction spans. Additionally, in the \textit{Proactive Alignment} category, MemoryOS achieves the highest QA score ($0.571$), indicating a superior capability in anticipating future steps and aligning with the user's long-term objectives.

\subsection{Scenario-Based Performance Analysis}

\begin{wrapfigure}{r}{0.40\textwidth} 
  \vspace{-10pt} 
  \centering
  \includegraphics[width=\linewidth]{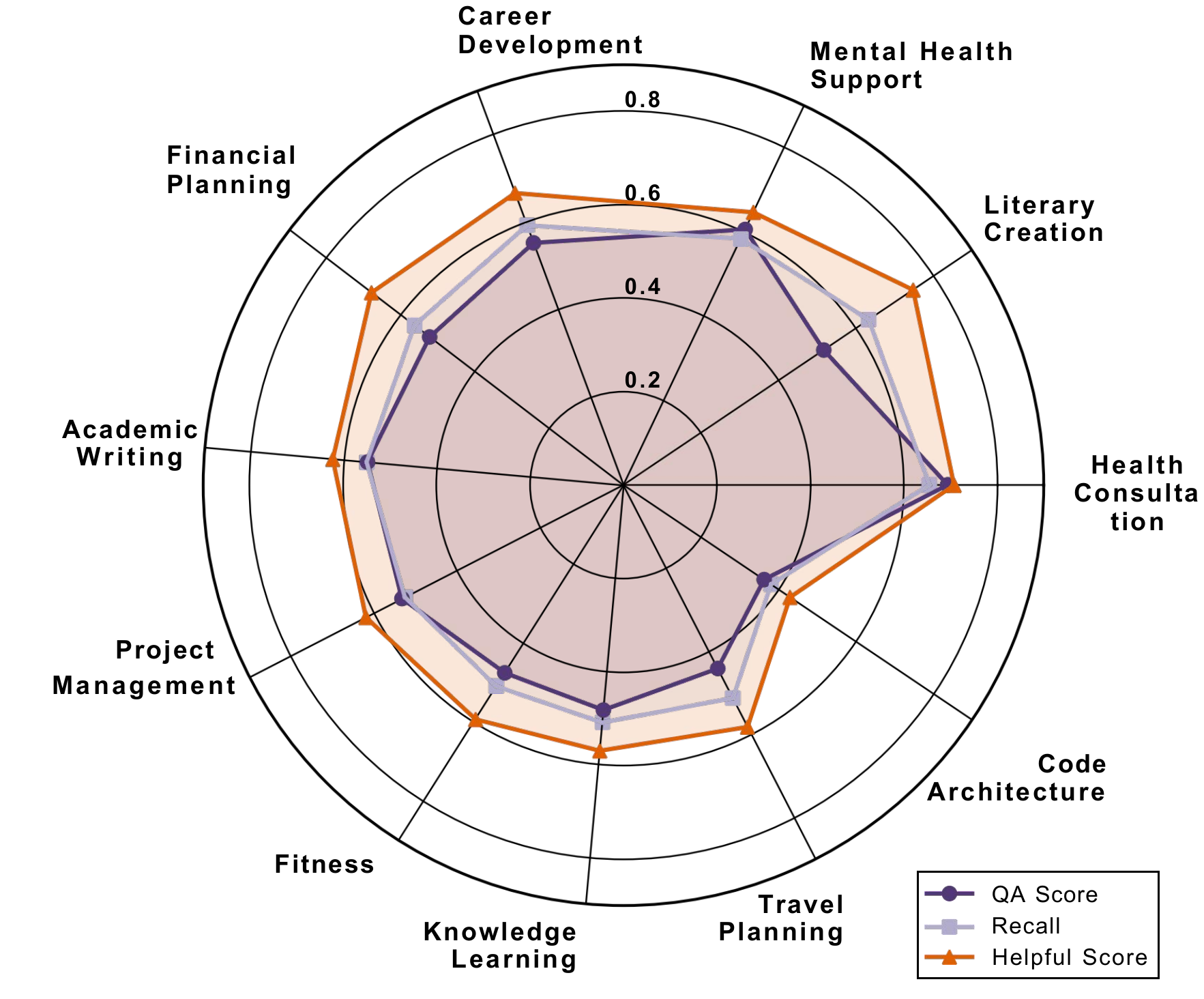}
  \vspace{-10pt}
  \caption{Average performance scores of MemoryOS across various topics.}
  \label{fig:topic}
\end{wrapfigure}

In this section, we present the memory system performance across various topics. To visualize trends, we select \textit{MemoryOS}, the most robust model, as the representative. Notably, other baselines exhibit similar tendencies; detailed results are provided in the Appendix.

As illustrated in Figure~\ref{fig:topic}, the efficacy of MemoryOS reveals significant domain-dependent variations. The system exhibits its strongest performance in consultative and creative domains, such as \textit{Mental Health Support}, \textit{Health Consultation}, and \textit{Literary Creation}. In these dimensions, the \textit{Helpful Score} (indicated by the orange trajectory) consistently outperforms other metrics, approaching or exceeding the $0.6$ to $0.8$ range. This trend suggests that for open-ended, human-centric tasks, the system is capable of providing highly valuable feedback and maintaining a coherent narrative flow, even when precise \textit{Recall} is not optimal.

Conversely, the system encounters severe challenges in highly technical and rigid domains. This is most distinct in \textit{Code Architecture}, where all three metrics—\textit{QA Score}, \textit{Recall}, and \textit{Helpful Score}—show a sharp contraction towards the center of the chart, dropping significantly below $0.4$. Unlike soft-constraint tasks like \textit{Financial Planning}, the rigorous dependency tracking and strict logical consistency required for complex engineering tasks expose a structural limitation in the current architecture's ability to handle hard constraints effectively.

\subsection{Efficiency and Cost Analysis of Long-Term Memory Systems}
\begin{table}[htbp]
\centering
\begin{tabular}{lccc}
\toprule
\textbf{Method} 
& \textbf{Avg. Add Memory Time (s)} 
& \textbf{Avg. Retrieve Memory Time (s)} 
& \textbf{Cost Token} \\
\midrule
Graph Mem & 26.898 & 0.18803 & 1224.02 \\
Mem0      & 11.751 & 1.261   & 4178.18 \\
A-mem     & 8.663  & 0.013   & 500.32  \\
MemoryOS  & 14.755 & 0.151   & 9201.01 \\
\bottomrule
\end{tabular}
\caption{Comparison of efficiency and cost across different memory systems. We report the average latency for memory addition and retrieval, alongside total token consumption.}
\label{tab:memory_performance}
\end{table}

Table~\ref{tab:memory_performance} reveals a critical bottleneck: memory incorporation latency consistently exceeds retrieval speeds across all systems. This underscores the urgent need to optimize memory ingestion mechanisms for scalable deployment.

Our evaluation of system latency and token consumption reveals a significant trade-off between operational overhead and model performance. Specifically, while MemoryOS achieves competitive retrieval latency (0.151\,s), it incurs the highest token cost (9,201), indicating that its superior performance necessitates extensive context maintenance. Conversely, A-mem demonstrates the lowest latency and resource consumption, yet as noted previously, this efficiency comes at the expense of retrieval accuracy and response quality. Graph Memory offers a viable compromise by delivering retrieval speeds comparable to MemoryOS (0.188\,s vs. 0.151\,s) with significantly reduced token costs, although it requires longer processing times for memory addition.

\subsection{Cross-Validation by Humans}

\begin{table}[htbp]
\centering
\renewcommand{\arraystretch}{1.2}
\begin{tabular}{lccc}
\toprule
\textbf{System} & \textbf{Avg. QA Score} & \textbf{QA Rank} & \textbf{Avg. Human Rank} \\
\midrule
MemoryOS  & \textbf{1.40} & \textbf{1} & \textbf{1.60} \\
Graph Mem & 1.10 & 2 & 2.20 \\
A-mem     & 1.00 & 3 & 2.60 \\
Mem0      & 0.80 & 4 & 3.60 \\
\bottomrule
\end{tabular}
\caption{Comparison of automated QA scores and human evaluation. The results demonstrate a strong alignment between our QA metrics and human judgments, with \textbf{MemoryOS} ranking first in both settings.}
\label{tab:human_qa_comparison}
\end{table}

Table \ref{tab:human_qa_comparison} presents the comparative results of average QA scores and human rankings. MemoryOS demonstrates superior performance, achieving the highest QA score ($1.40$) and the best average human ranking ($1.60$). Crucially, the ranking order remains consistent across all evaluated systems $(\text{MemoryOS} > \text{Graph Memory} > \text{A-mem} > \text{Mem0})$, where a decrease in QA score strictly corresponds to a decline in human ranking (e.g., Mem0 scores lowest at $0.80$ and ranks last at $3.60$). This perfect alignment in system-level ranking further validates that our automated QA Score serves as a reliable indicator of model performance consistent with human preference.
\section{Conclusion}

We introduce \textbf{RealMem}, a benchmark for evaluating long-term memory in realistic, project-oriented interactions, where agents must track evolving goals and maintain coherent project states across sessions. RealMem departs from prior benchmarks by emphasizing sustained, multi-session projects and by providing a synthesis pipeline that enables controlled simulation of dynamic memory evolution.
Experiments on RealMem show that existing memory systems remain fundamentally limited in handling long-term project dependencies. 
The substantial gap between all evaluated methods and the Oracle upper bound further highlights that effective long-term memory integration remains a core bottleneck for project-oriented agents.
We believe RealMem provides a valuable diagnostic testbed for studying long-term agent memory and hope it will facilitate future research toward more robust, scalable, and reliable memory systems for real-world autonomous agents.


\section*{Limitations}

Our data construction process relies significantly on the Gemini 2.5 series models for data collection, alongside human annotation for label verification. While this reliance may raise concerns regarding reproducibility and associated costs, the Gemini 2.5 models demonstrate a superior capability to simulate realistic human--computer interactions with highly controllable memory utilization. In contrast, our experiments with the GPT series revealed inferior format adherence, leading us to ultimately select the Gemini models. In terms of evaluation scope, RealMem currently focuses specifically on memory-centric challenges within long-term project-oriented interactions. Consequently, the benchmark does not yet involve the assessment of tool use capabilities, which we plan to incorporate in future work to further extend its applicability in complex task processing.

\bibliography{custom}

@article{maharana2024evaluating,
  title={Evaluating very long-term conversational memory of llm agents},
  author={Maharana, Adyasha and Lee, Dong-Ho and Tulyakov, Sergey and Bansal, Mohit and Barbieri, Francesco and Fang, Yuwei},
  journal={arXiv preprint arXiv:2402.17753},
  year={2024}
}

@article{wu2024longmemeval,
  title={Longmemeval: Benchmarking chat assistants on long-term interactive memory},
  author={Wu, Di and Wang, Hongwei and Yu, Wenhao and Zhang, Yuwei and Chang, Kai-Wei and Yu, Dong},
  journal={arXiv preprint arXiv:2410.10813},
  year={2024}
}

@article{chen2025halumem,
  title={HaluMem: Evaluating Hallucinations in Memory Systems of Agents},
  author={Chen, Ding and Niu, Simin and Li, Kehang and Liu, Peng and Zheng, Xiangping and Tang, Bo and Li, Xinchi and Xiong, Feiyu and Li, Zhiyu},
  journal={arXiv preprint arXiv:2511.03506},
  year={2025}
}

@article{jiang2025personamem,
  title={PersonaMem-v2: Towards Personalized Intelligence via Learning Implicit User Personas and Agentic Memory},
  author={Jiang, Bowen and Yuan, Yuan and Shen, Maohao and Hao, Zhuoqun and Xu, Zhangchen and Chen, Zichen and Liu, Ziyi and Vijjini, Anvesh Rao and He, Jiashu and Yu, Hanchao and others},
  journal={arXiv preprint arXiv:2512.06688},
  year={2025}
}

@article{wei2025evo,
  title={Evo-Memory: Benchmarking LLM Agent Test-time Learning with Self-Evolving Memory},
  author={Wei, Tianxin and Sachdeva, Noveen and Coleman, Benjamin and He, Zhankui and Bei, Yuanchen and Ning, Xuying and Ai, Mengting and Li, Yunzhe and He, Jingrui and Chi, Ed H and others},
  journal={arXiv preprint arXiv:2511.20857},
  year={2025}
}

@article{gutierrez2025rag,
  title={From rag to memory: Non-parametric continual learning for large language models},
  author={Guti{\'e}rrez, Bernal Jim{\'e}nez and Shu, Yiheng and Qi, Weijian and Zhou, Sizhe and Su, Yu},
  journal={arXiv preprint arXiv:2502.14802},
  year={2025}
}

@article{Thang2025marag,
  title={MA-RAG: Multi-Agent Retrieval-Augmented Generation via Collaborative Chain-of-Thought Reasoning},
  author={Thang, Nguyen and Chin, Peter and Nguyen, Thang and Tai, Yu-Wing and Chin, Peter and Tai, Yu-Wing},
  year={2025},
  journal={http://arxiv.org/abs/2505.20096},
  volume={abs/2505.20096},
  doi={10.48550/arxiv.2505.20096},
}

@article{chhikara2025mem0,
  title={Mem0: Building production-ready ai agents with scalable long-term memory},
  author={Chhikara, Prateek and Khant, Dev and Aryan, Saket and Singh, Taranjeet and Yadav, Deshraj},
  journal={arXiv preprint arXiv:2504.19413},
  year={2025}
}

@article{xu2025mem,
  title={A-mem: Agentic memory for llm agents},
  author={Xu, Wujiang and Liang, Zujie and Mei, Kai and Gao, Hang and Tan, Juntao and Zhang, Yongfeng},
  journal={arXiv preprint arXiv:2502.12110},
  year={2025}
}

@article{rasmussen2025zep,
  title={Zep: a temporal knowledge graph architecture for agent memory},
  author={Rasmussen, Preston and Paliychuk, Pavlo and Beauvais, Travis and Ryan, Jack and Chalef, Daniel},
  journal={arXiv preprint arXiv:2501.13956},
  year={2025}
}

@misc{kang2025memoryosaiagent,
      title={Memory OS of AI Agent}, 
      author={Jiazheng Kang and Mingming Ji and Zhe Zhao and Ting Bai},
      year={2025},
      eprint={2506.06326},
      archivePrefix={arXiv},
      primaryClass={cs.AI},
      url={https://arxiv.org/abs/2506.06326}, 
}

@article{fang2025lightmem,
  title={Lightmem: Lightweight and efficient memory-augmented generation},
  author={Fang, Jizhan and Deng, Xinle and Xu, Haoming and Jiang, Ziyan and Tang, Yuqi and Xu, Ziwen and Deng, Shumin and Yao, Yunzhi and Wang, Mengru and Qiao, Shuofei and others},
  journal={arXiv preprint arXiv:2510.18866},
  year={2025}
}

@article{li2025memos,
  title={Memos: A memory os for ai system},
  author={Li, Zhiyu and Song, Shichao and Xi, Chenyang and Wang, Hanyu and Tang, Chen and Niu, Simin and Chen, Ding and Yang, Jiawei and Li, Chunyu and Yu, Qingchen and others},
  journal={arXiv preprint arXiv:2507.03724},
  year={2025}
}

@misc{wang2025mirixmultiagentmemoryllmbased,
      title={MIRIX: Multi-Agent Memory System for LLM-Based Agents}, 
      author={Yu Wang and Xi Chen},
      year={2025},
      eprint={2507.07957},
      archivePrefix={arXiv},
      primaryClass={cs.CL},
      url={https://arxiv.org/abs/2507.07957}, 
}

@article{tran2025multi,
  title={Multi-agent collaboration mechanisms: A survey of llms},
  author={Tran, Khanh-Tung and Dao, Dung and Nguyen, Minh-Duong and Pham, Quoc-Viet and O'Sullivan, Barry and Nguyen, Hoang D},
  journal={arXiv preprint arXiv:2501.06322},
  year={2025}
}

@article{sapkota2025ai,
  title={Ai agents vs. agentic ai: A conceptual taxonomy, applications and challenges},
  author={Sapkota, Ranjan and Roumeliotis, Konstantinos I and Karkee, Manoj},
  journal={arXiv preprint arXiv:2505.10468},
  year={2025}
}

@article{hu2025memory,
  title={Memory in the Age of AI Agents},
  author={Hu, Yuyang and Liu, Shichun and Yue, Yanwei and Zhang, Guibin and Liu, Boyang and Zhu, Fangyi and Lin, Jiahang and Guo, Honglin and Dou, Shihan and Xi, Zhiheng and others},
  journal={arXiv preprint arXiv:2512.13564},
  year={2025}
}

@article{tang2025agent,
  title={Agent kb: Leveraging cross-domain experience for agentic problem solving},
  author={Tang, Xiangru and Qin, Tianrui and Peng, Tianhao and Zhou, Ziyang and Shao, Daniel and Du, Tingting and Wei, Xinming and Xia, Peng and Wu, Fang and Zhu, He and others},
  journal={arXiv preprint arXiv:2507.06229},
  year={2025}
}

@article{fang2025comprehensive,
  title={A comprehensive survey of self-evolving ai agents: A new paradigm bridging foundation models and lifelong agentic systems},
  author={Fang, Jinyuan and Peng, Yanwen and Zhang, Xi and Wang, Yingxu and Yi, Xinhao and Zhang, Guibin and Xu, Yi and Wu, Bin and Liu, Siwei and Li, Zihao and others},
  journal={arXiv preprint arXiv:2508.07407},
  year={2025}
}

@article{liu2024toolace,
  title={Toolace: Winning the points of llm function calling},
  author={Liu, Weiwen and Huang, Xu and Zeng, Xingshan and Hao, Xinlong and Yu, Shuai and Li, Dexun and Wang, Shuai and Gan, Weinan and Liu, Zhengying and Yu, Yuanqing and others},
  journal={arXiv preprint arXiv:2409.00920},
  year={2024}
}

@article{prabhakar2025apigen,
  title={Apigen-mt: Agentic pipeline for multi-turn data generation via simulated agent-human interplay},
  author={Prabhakar, Akshara and Liu, Zuxin and Zhu, Ming and Zhang, Jianguo and Awalgaonkar, Tulika and Wang, Shiyu and Liu, Zhiwei and Chen, Haolin and Hoang, Thai and Niebles, Juan Carlos and others},
  journal={arXiv preprint arXiv:2504.03601},
  year={2025}
}

@article{xia2025storywriter,
  title={StoryWriter: A Multi-Agent Framework for Long Story Generation},
  author={Xia, Haotian and Peng, Hao and Qi, Yunjia and Wang, Xiaozhi and Xu, Bin and Hou, Lei and Li, Juanzi},
  journal={arXiv preprint arXiv:2506.16445},
  year={2025}
}

@article{huang2024agentcoder,
  title={AgentCoder: Multi-Agent Code Generation with Effective Testing and Self-Optimization},
  author={Huang, D and Zhang, JM and Luck, M and Bu, Q and Qing, Y and Cui, H},
  journal={University of Hong Kong, King’s College London, University of Sussex, Shanghai Jiao Tong University},
  year={2024}
}

@article{levi2025intellagent,
  title={Intellagent: A multi-agent framework for evaluating conversational ai systems},
  author={Levi, Elad and Kadar, Ilan},
  journal={arXiv preprint arXiv:2501.11067},
  year={2025}
}

@online{openai_text_embedding_3_small_docs,
  author       = {{OpenAI}},
  title        = {text-embedding-3-small (Model documentation)},
  year         = {2024},
  url          = {https://platform.openai.com/docs/models/text-embedding-3-small},
}

@article{ouyang2025reasoningbank,
  title={Reasoningbank: Scaling agent self-evolving with reasoning memory},
  author={Ouyang, Siru and Yan, Jun and Hsu, I and Chen, Yanfei and Jiang, Ke and Wang, Zifeng and Han, Rujun and Le, Long T and Daruki, Samira and Tang, Xiangru and others},
  journal={arXiv preprint arXiv:2509.25140},
  year={2025}
}

@inproceedings{wang2024recmind,
  title={Recmind: Large language model powered agent for recommendation},
  author={Wang, Yancheng and Jiang, Ziyan and Chen, Zheng and Yang, Fan and Zhou, Yingxue and Cho, Eunah and Fan, Xing and Lu, Yanbin and Huang, Xiaojiang and Yang, Yingzhen},
  booktitle={Findings of the Association for Computational Linguistics: NAACL 2024},
  pages={4351--4364},
  year={2024}
}

@misc{kim2025principlessyntheticstrategymemory,
      title={PRINCIPLES: Synthetic Strategy Memory for Proactive Dialogue Agents}, 
      author={Namyoung Kim and Kai Tzu-iunn Ong and Yeonjun Hwang and Minseok Kang and Iiseo Jihn and Gayoung Kim and Minju Kim and Jinyoung Yeo},
      year={2025},
      eprint={2509.17459},
      archivePrefix={arXiv},
      primaryClass={cs.CL},
      url={https://arxiv.org/abs/2509.17459}, 
}

@article{liu2023g,
  title={G-eval: NLG evaluation using gpt-4 with better human alignment},
  author={Liu, Yang and Iter, Dan and Xu, Yichong and Wang, Shuohang and Xu, Ruochen and Zhu, Chenguang},
  journal={arXiv preprint arXiv:2303.16634},
  year={2023}
}

@article{hurst2024gpt,
  title={Gpt-4o system card},
  author={Hurst, Aaron and Lerer, Adam and Goucher, Adam P and Perelman, Adam and Ramesh, Aditya and Clark, Aidan and Ostrow, AJ and Welihinda, Akila and Hayes, Alan and Radford, Alec and others},
  journal={arXiv preprint arXiv:2410.21276},
  year={2024}
}

@inproceedings{tang2025synthesizing,
  title={Synthesizing post-training data for llms through multi-agent simulation},
  author={Tang, Shuo and Pang, Xianghe and Liu, Zexi and Tang, Bohan and Ye, Rui and Jin, Tian and Dong, Xiaowen and Wang, Yanfeng and Chen, Siheng},
  booktitle={Proceedings of the 63rd Annual Meeting of the Association for Computational Linguistics (Volume 1: Long Papers)},
  pages={23306--23335},
  year={2025}
}

@inproceedings{zhang2024multimodal,
  title={A multimodal foundation agent for financial trading: Tool-augmented, diversified, and generalist},
  author={Zhang, Wentao and Zhao, Lingxuan and Xia, Haochong and Sun, Shuo and Sun, Jiaze and Qin, Molei and Li, Xinyi and Zhao, Yuqing and Zhao, Yilei and Cai, Xinyu and others},
  booktitle={Proceedings of the 30th acm sigkdd conference on knowledge discovery and data mining},
  pages={4314--4325},
  year={2024}
}

@inproceedings{li2025hello,
  title={Hello again! llm-powered personalized agent for long-term dialogue},
  author={Li, Hao and Yang, Chenghao and Zhang, An and Deng, Yang and Wang, Xiang and Chua, Tat-Seng},
  booktitle={Proceedings of the 2025 Conference of the Nations of the Americas Chapter of the Association for Computational Linguistics: Human Language Technologies (Volume 1: Long Papers)},
  pages={5259--5276},
  year={2025}
}

@article{majumder2023clin,
  title={Clin: A continually learning language agent for rapid task adaptation and generalization},
  author={Majumder, Bodhisattwa Prasad and Mishra, Bhavana Dalvi and Jansen, Peter and Tafjord, Oyvind and Tandon, Niket and Zhang, Li and Callison-Burch, Chris and Clark, Peter},
  journal={arXiv preprint arXiv:2310.10134},
  year={2023}
}

@article{zhang2025agentic,
  title={Agentic context engineering: Evolving contexts for self-improving language models},
  author={Zhang, Qizheng and Hu, Changran and Upasani, Shubhangi and Ma, Boyuan and Hong, Fenglu and Kamanuru, Vamsidhar and Rainton, Jay and Wu, Chen and Ji, Mengmeng and Li, Hanchen and others},
  journal={arXiv preprint arXiv:2510.04618},
  year={2025}
}

@article{zhang2025survey,
  title={A survey on the memory mechanism of large language model-based agents},
  author={Zhang, Zeyu and Dai, Quanyu and Bo, Xiaohe and Ma, Chen and Li, Rui and Chen, Xu and Zhu, Jieming and Dong, Zhenhua and Wen, Ji-Rong},
  journal={ACM Transactions on Information Systems},
  volume={43},
  number={6},
  pages={1--47},
  year={2025},
  publisher={ACM New York, NY}
}

@misc{hu2026doesmemoryneedgraphs,
      title={Does Memory Need Graphs? A Unified Framework and Empirical Analysis for Long-Term Dialog Memory}, 
      author={Sen Hu and Yuxiang Wei and Jiaxin Ran and Zhiyuan Yao and Lei Zou},
      year={2026},
      eprint={2601.01280},
      archivePrefix={arXiv},
      primaryClass={cs.CL},
      url={https://arxiv.org/abs/2601.01280}, 
}

@inproceedings{du2025graphmaster,
  author    = {Du, Enjun and Li, Xunkai and Jin, Tian and Zhang, Zhihan and Li, Rong{-}Hua and Wang, Guoren},
  title     = {GraphMaster: Automated Graph Synthesis via {LLM} Agents in Data-Limited Environments},
  booktitle = {Advances in Neural Information Processing Systems 39 (NeurIPS 2025)},
  year      = {2025},
  month     = apr
}

@article{du2025graphoracle,
  title={GraphOracle: A Foundation Model for Knowledge Graph Reasoning},
  author={Du, Enjun and Liu, Siyu and Zhang, Yongqi},
  journal={arXiv preprint arXiv:2505.11125},
  year={2025},
  url={https://arxiv.org/abs/2505.11125}
}

@misc{guo2025octopusagenticmultimodalreasoning,
      title={Octopus: Agentic Multimodal Reasoning with Six-Capability Orchestration}, 
      author={Yifu Guo and Zishan Xu and Zhiyuan Yao and Yuquan Lu and Jiaye Lin and Sen Hu and Zhenheng Tang and Huacan Wang and Ronghao Chen},
      year={2025},
      eprint={2511.15351},
      archivePrefix={arXiv},
      primaryClass={cs.AI},
      url={https://arxiv.org/abs/2511.15351}, 
}

@misc{luo2025codetestcasesenough,
      title={How Many Code and Test Cases Are Enough? Evaluating Test Cases Generation from a Binary-Matrix Perspective}, 
      author={Xianzhen Luo and Jinyang Huang and Wenzhen Zheng and Qingfu Zhu and Mingzheng Xu and Yiheng Xu and Yuantao Fan and Libo Qin and Wanxiang Che},
      year={2025},
      eprint={2510.08720},
      archivePrefix={arXiv},
      primaryClass={cs.CL},
      url={https://arxiv.org/abs/2510.08720}, 
}
\clearpage

\appendix
\section{Appendix}
\subsection{Supplementary Details of RealMem}

This appendix provides additional statistical information and key definitions of the RealMem dataset to support a more detailed understanding of its data composition and task taxonomy.

\begin{table}[htbp]
\centering
\begin{tabular}{l r}
\toprule
\textbf{Statistic} & \textbf{Value} \\
\midrule
\multicolumn{2}{l}{\textit{Dialogue Scale}} \\
Avg. Context Length (tokens / user) & 269{,}190 \\
Avg. Session Number (per user)      & 205 \\
Avg. Turns per Session              & 6.8 \\
Total Dialogue Turns                & 14{,}028 \\
\midrule
\multicolumn{2}{l}{\textit{Memory Composition}} \\
Total Memories                      & 5{,}072 \\
\quad Project-centric Event         & 4{,}135 \\
\quad Persona                       & 850 \\
\quad Schedule                      & 87 \\
Used Memory (Ground Truth)          & 1{,}834 \\
\midrule
\multicolumn{2}{l}{\textit{Question Distribution}} \\
Total Questions                     & 1{,}415 \\
\quad Static Retrieval              & 1{,}075 \\
\quad Dynamic Updating              & 156 \\
\quad Implicit Preference           & 160 \\
\quad Temporal Reasoning            & 24 \\
\bottomrule
\end{tabular}
\caption{Statistical overview of the RealMem dataset.}
\label{tab:dataset-statistics}
\end{table}

\subsubsection{Definition of Question Types}
To systematically evaluate the capabilities of our memory mechanism, we categorize user queries into four distinct types based on their retrieval requirements and interaction logic:

\begin{itemize}
    \item \textbf{Static Retrieval:} Queries that seek to advance the project state linearly based on confirmed context. In these scenarios, the user accepts the current progress (e.g., "The suggestions are great") and explicitly requests the subsequent step (e.g., "What do we need to do next?"). The agent must retrieve the latest stable state to build upon it without altering previous decisions.
    
    \item \textbf{Dynamic Updating:} Queries involving the modification, deletion, or conflict resolution of existing plans. Unlike incremental queries, the user introduces new constraints (e.g., "maintain 12 days but add West Coast") that conflict with established memories. The agent must retrieve relevant constraints and perform trade-off analysis to generate a revised state.
    
    \item \textbf{Proactive Alignment:} Queries lacking explicit instructions, often manifested as emotional feedback or vague statements (e.g., "This is fantastic!"). Here, the agent cannot rely on current turn commands but must retrieve long-term user priorities or historical preferences to proactively propose the next logical action, transforming latent intent into execution.
    
    \item \textbf{Temporal-reasoning:} Queries requiring the processing of time-sensitive information and scheduling logic. These tasks involve validating proposed slots against existing commitments (e.g., checking for overlaps between a new workout plan and existing study sessions) or sequencing events chronologically. The agent must retrieve structured schedule data to perform constraint satisfaction checks.
\end{itemize}

\subsubsection{Definition of Memory Types}
To support effective retrieval and distinct reasoning tasks, we structure the agent's memory into three specific categories:

\begin{table*}[t]
\centering
\small
\renewcommand{\arraystretch}{1.3} 

\begin{tabularx}{\linewidth}{>{\raggedright\arraybackslash}X} 
\toprule

\rowcolor{headergray} 
\textbf{Academic Writing} \\
Topic ideation, literature synthesis, structural outlining, argumentation refinement, and stylistic editing. \\

\rowcolor{headergray}
\textbf{Career Development} \\
CV/Resume polishing, mock interview simulation, behavioral question strategy, and career narrative refinement. \\

\rowcolor{headergray}
\textbf{Code Architecture Design} \\
System modularization, design pattern application, API interface definition, technical stack selection, and scalability planning. \\

\rowcolor{headergray}
\textbf{Financial Planning} \\
Long-term goal formulation, asset allocation, risk assessment, and financial literacy education. \\

\rowcolor{headergray}
\textbf{Fitness} \\
Personalized workout routines, dietary tracking, biometric monitoring, and progressive plan adaptation. \\

\rowcolor{headergray}
\textbf{Health Consultation} \\
Symptom triage, medical report interpretation, lifestyle health advisory, and chronic condition management. \\

\rowcolor{headergray}
\textbf{Knowledge Learning} \\
Curriculum scaffolding, progress tracking, knowledge reinforcement, and adaptive learning path design. \\

\rowcolor{headergray}
\textbf{Literary Creation} \\
Narrative ideation, world-building coherence, character development, and stylistic enhancement. \\

\rowcolor{headergray}
\textbf{Mental Health Support} \\
Emotional support, cognitive reframing strategies, interpersonal conflict resolution, and stress management. \\

\rowcolor{headergray}
\textbf{Project Management} \\
Task decomposition, resource scheduling, documentation maintenance, and retrospective analysis. \\

\rowcolor{headergray}
\textbf{Travel Planning} \\
Multi-day itinerary scheduling, POI recommendations, culinary integration, and dynamic constraint-based adjustments. \\

\bottomrule
\end{tabularx}
\caption{Overview of the 11 evaluation scenarios and their core task attributes in \textsc{RealMem}.}
\label{tab:scenarios_attributes}
\end{table*}

\begin{itemize}
    \item \textbf{Persona:} This category stores static and semi-static user attributes, including personal profiles, long-term goals, and specific preferences (e.g., dietary restrictions, preferred writing styles). It serves as the foundation for personalization, ensuring that the agent's responses consistently align with the user's identity and historical habits across different sessions.
    
    \item \textbf{Project States:} This category records the evolving state of the specific task or project. It encapsulates the core content generated during the collaboration, such as finalized itinerary details, code architecture decisions, or plot outlines. These memories act as the "knowledge base" for the project, allowing the agent to track progress and maintain context continuity without redundant inquiries.
    
    \item \textbf{Schedule:} This category contains structured, time-sensitive data representing the user's global timeline. It includes explicit appointments, deadlines, and recurring routines (e.g., "Gym every Tuesday at 6 PM"). The primary function of this memory type is to facilitate temporal reasoning and conflict detection, ensuring that proposed project plans do not violate existing temporal constraints.
\end{itemize}

\subsubsection{Definition of Topic Types}

To ensure the dataset reflects the diversity and complexity of real-world human-AI interactions, we curated eleven high-frequency user scenarios. 
As detailed in Table~\ref{tab:scenarios_attributes}, these scenarios span multiple distinct domains:
\begin{itemize}
    \item \textbf{Life Planning \& Management:} Scenarios like \textit{Travel Planning}, \textit{Financial Planning}, and \textit{Project Management} that require the agent to handle dynamic scheduling, resource allocation, and hierarchical goal decomposition.
    \item \textbf{Professional \& Skill Development:} Tasks such as \textit{Academic Writing}, \textit{Literary Creation}, \textit{Career Development}, \textit{Knowledge Learning}, and \textit{Code Architecture Design}, focusing on iterative content refinement, technical precision, and long-term learning trajectories.
    \item \textbf{Personal Well-being:} Sensitive domains including \textit{Fitness}, \textit{Mental Health Support}, and \textit{Health Consultation}, which demand high personalization, empathy, and strict adherence to context-specific constraints.
\end{itemize}
For each scenario, we explicitly modeled core attributes—ranging from budget constraints to emotional coping mechanisms—to establish clear evaluation boundaries.

\begin{table*}[h]
\centering
\renewcommand{\arraystretch}{1.15} 

\resizebox{\textwidth}{!}{%
\begin{tabular}{l cc cc cc cc}
\toprule
\multirow{2}{*}{Topic}
& \multicolumn{2}{c}{A-Mem}
& \multicolumn{2}{c}{Mem0}
& \multicolumn{2}{c}{MemoryOS}
& \multicolumn{2}{c}{Graph Memory} \\
\cmidrule(lr){2-3}\cmidrule(lr){4-5}\cmidrule(lr){6-7}\cmidrule(lr){8-9}
& QA & Recall & QA & Recall & QA & Recall & QA & Recall \\
\midrule
Academic Writing         & 0.383 & 0.504 & 0.517 & 0.554 & 0.550 & 0.552 & 0.450 & 0.458 \\
Career Development       & 0.485 & 0.517 & 0.515 & 0.502 & 0.553 & 0.593 & 0.492 & 0.467 \\
Code Architecture Design & \underline{0.322} & \underline{0.291} & \underline{0.380} & \underline{0.446} & \underline{0.363} & \underline{0.381} & \underline{0.363} & \underline{0.345} \\
Financial Planning       & 0.453 & \textbf{0.532} & 0.453 & 0.540 & 0.522 & 0.563 & 0.497 & 0.577 \\
Fitness                  & 0.418 & 0.423 & 0.482 & 0.548 & 0.476 & 0.510 & 0.503 & 0.489 \\
Health Consultation      & \textbf{0.556} & 0.470 & \textbf{0.681} & 0.569 & \textbf{0.694} & \textbf{0.655} & \textbf{0.681} & \textbf{0.651} \\
Knowledge Learning       & 0.432 & 0.438 & 0.445 & 0.519 & 0.484 & 0.511 & 0.482 & 0.481 \\
Literary Creation        & 0.437 & 0.516 & 0.498 & \textbf{0.581} & 0.517 & 0.633 & 0.502 & 0.532 \\
Mental Health Support    & 0.465 & 0.441 & 0.511 & 0.526 & 0.606 & 0.584 & 0.569 & 0.552 \\
Project Management       & 0.420 & 0.506 & 0.447 & 0.488 & 0.533 & 0.526 & 0.453 & 0.438 \\
Travel Planning          & 0.368 & 0.462 & 0.385 & 0.534 & 0.442 & 0.513 & 0.424 & 0.483 \\
\bottomrule
\end{tabular}
}
\caption{Performance variability across topics. For each method, the best-performing topic is highlighted in \textbf{bold}, and the worst-performing topic is \underline{underlined}. Comparisons are made column-wise (per method).}
\label{tab:all-topic-results}
\end{table*}

\subsection{Comparison of QA Score and Recall across different topics for four methods.}

Table~\ref{tab:all-topic-results} reveals a significant domain sensitivity across all evaluated architectures, characterized by a sharp performance stratification between rigid and entity-centric tasks. Specifically, \textit{Code Architecture Design} consistently yields the lowest scores across all methods (e.g., MemoryOS QA: $0.363$), indicating that current retrieval-based memory systems struggle with engineering tasks demanding strict logical consistency and precise dependency tracking, where fuzzy semantic matching is often insufficient. In contrast, \textit{Health Consultation} emerges as the most robust scenario (e.g., MemoryOS QA: $0.694$), likely because medical dialogues revolve around distinct named entities (e.g., symptoms, medications) that serve as strong semantic anchors for retrieval. This divergence suggests that while current long-term memory paradigms are effective for open-ended, consultative interactions, they remain insufficient for problem-solving scenarios that require high-precision state maintenance.

\subsection{Case Studies}

To provide a clearer illustration of our defined problem types, we present the following case studies. These examples serve to concretize the abstract definitions, demonstrating how the agent must leverage distinct memory operations—ranging from incremental reasoning to conflict resolution—to address specific challenges in long-term interactions.

As illustrated in the case study below, the user acknowledges prior context (attractions and food suggestions) and explicitly initiates a state transition toward itinerary planning. The system leverages memory to confirm these preferences and proposes a logical continuation. This exemplifies \textit{Dynamic Incremental (Explicit Progression)}, where the dialogue builds constructively upon established history to advance the specific task goal.

\begin{tcolorbox}[
    enhanced,
    breakable,                
    colframe=black!70,        
    colback=gray!5,           
    coltitle=white,           
    fonttitle=\bfseries,      
    title=Dynamic Incremental,
    arc=2pt,                  
    boxrule=0.8pt,            
    left=6pt, right=6pt, top=6pt, bottom=6pt 
]
    \footnotesize
    $\blacktriangleright$ \textbf{User:}\\
    Hi! I've had a look at the updated attractions list. The foodie suggestions are great! I'm excited to start mapping out the actual day-by-day itinerary now. What do we need to do next?
    
    \vspace{0.5em}
    $\blacktriangleright$ \textbf{Assistant:}\\
    That's fantastic to hear! I'm glad you liked the foodie suggestions.
    
    Since we have the overall 'Classic Loop' itinerary and now a great list of local food experiences, the next logical step is exactly what you said: mapping out the actual day-by-day itinerary, making sure to weave in those delicious foodie stops.
    
    Would you like to start by reviewing Day 1 and Day 2, or perhaps focus on specific regions first, like Queenstown or the West Coast, to integrate the food elements?
    
    \vspace{0.5em}
    \hrule \vspace{0.5em} 
    
    $\blacktriangleright$ \textbf{Memory Used:}\\
    \texttt{[Food Scheme]} $\rightarrow$ \texttt{[Application]}: Acknowledge user's positive feedback on foodie suggestions and propose integrating them into the day-by-day itinerary.
\end{tcolorbox}

\vspace{1em}
As illustrated in the case study below, the user's input is predominantly emotional and affirmative, lacking explicit directives or queries. To maintain dialogue momentum, the assistant proactively retrieves a stored long-term priority (booking flights and RV rental) to propose the next logical step. This exemplifies \textit{Proactive Alignment}, where the system utilizes memory to bridge the conversational gap and guide the planning flow when immediate user intent is ambiguous.

\begin{tcolorbox}[
    enhanced,
    breakable,
    colframe=black!70,
    colback=gray!5,
    fonttitle=\bfseries,
    title=Proactive Alignment,
    arc=2pt, boxrule=0.8pt,
    left=6pt, right=6pt, top=6pt, bottom=6pt
]
    \footnotesize
    $\blacktriangleright$ \textbf{User:}\\
    Wow, this is fantastic! You've really outdone yourself. The foodie additions are exactly what I was hoping for, and you've hit so many great spots. Fergburger is definitely going straight to the top of my list, and local salmon in Tekapo sounds absolutely amazing. I'm already imagining all the delicious food!
    
    This enhanced guide sounds absolutely perfect, and yes, my taste buds are definitely ready to be tantalized!
    
    \vspace{0.5em}
    $\blacktriangleright$ \textbf{Assistant:}\\
    That's wonderful to hear! I'm so glad the foodie additions hit the mark and you're excited about all the delicious experiences awaiting you in New Zealand.
    
    Now that we have a fantastic baseline for attractions and a mouth-watering foodie guide, we can move on to the next crucial step. You previously mentioned that booking flights and the RV rental would be the most important next priority to lock in. Shall we dive into those details now?
    
    \vspace{0.5em}
    \hrule \vspace{0.5em}
    
    $\blacktriangleright$ \textbf{Memory Used:}\\
    \texttt{[Priority: Flights \& RV]} $\rightarrow$ \texttt{[Application]}: Propose the next step in trip planning, which is booking flights and RV rental, as confirmed by the user previously.
\end{tcolorbox}

\vspace{1em}
As illustrated in the case study below, the user introduces a constraint conflict: maintaining a fixed 12-day duration while integrating substantial new activities (West Coast). This necessitates a structural modification of the existing itinerary rather than a simple addition. The system retrieves specific plan details (e.g., Queenstown commitments) to evaluate the impact of resource reallocation. This exemplifies \textit{Dynamic Updating}, where the agent leverages memory to negotiate trade-offs and alter the plan's state to accommodate conflicting requirements.

\begin{tcolorbox}[
    enhanced,
    breakable,
    colframe=black!70,
    colback=gray!5,
    fonttitle=\bfseries,
    title=Dynamic Updating,
    arc=2pt, boxrule=0.8pt,
    left=6pt, right=6pt, top=6pt, bottom=6pt
]
    \footnotesize
    $\blacktriangleright$ \textbf{User:}\\
    That sounds amazing, I really want to do it! Hmm, extending the trip might be tough with my work schedule, so I'd really prefer to stick to 12 days if possible.
    
    But I also don't want the whole trip to feel super rushed and stressful, you know? Can you show me what it would look like if we tried to fit the two West Coast days into the existing 12 days? I want to see how much more 'rushed' it would actually be before I decide.
    
    \vspace{0.5em}
    $\blacktriangleright$ \textbf{Assistant:}\\
    I understand completely! Sticking to 12 days while adding two full West Coast days is a challenge, but we can definitely map it out so you can see the trade-offs.
    
    To incorporate two \textit{full} days on the West Coast for your heli-hike and exploration, we'd need to allocate three nights in the Franz Josef/Fox Glacier area instead of the original one night. This means we need to free up two days from elsewhere in the itinerary.
    
    The most practical way to do this without drastically altering the core...
    
    \vspace{0.5em}
    \hrule \vspace{0.5em}
    
    $\blacktriangleright$ \textbf{Memory Used:}\\
    \texttt{[Plan Detail: Queenstown Day 4]} $\rightarrow$ \texttt{[Application]}: Identify the existing commitments in Queenstown that would be impacted by reducing the time allocated there.
\end{tcolorbox}

\vspace{1em}
As illustrated in the case study below, the user proposes a recurring workout schedule for specific days. To validate feasibility, the system retrieves existing temporal constraints (Study Sessions) from memory. By cross-referencing the proposed days with occupied time slots to ensure no overlap, the agent performs constraint satisfaction. This exemplifies \textit{Temporal Reasoning}, where the system processes chronological relationships to maintain schedule consistency.

\begin{tcolorbox}[
    enhanced,
    breakable,
    colframe=black!70,
    colback=gray!5,
    fonttitle=\bfseries,
    title=Temporal Reasoning,
    arc=2pt, boxrule=0.8pt,
    left=6pt, right=6pt, top=6pt, bottom=6pt
]
    \footnotesize
    $\blacktriangleright$ \textbf{User:}\\
    Hello, I want to arrange a workout schedule for Tuesdays, Thursdays, and Saturdays. How about that?
    
    \vspace{0.5em}
    $\blacktriangleright$ \textbf{Assistant:}\\
    Yes, that sounds excellent! I've checked your proposed times against your study sessions, and they fit perfectly without any overlaps.
    
    So, to confirm, your fitness schedule will be:
    \begin{itemize}[leftmargin=1.5em, itemsep=0pt, parsep=0pt, topsep=2pt]
        \item \textbf{Tuesdays:} 6:00 PM - 6:45 PM
        \item \textbf{Thursdays:} 6:00 PM - 6:45 PM
        \item \textbf{Saturdays:} 9:00 AM - 9:45 AM
    \end{itemize}
    Now that we have the times locked in, the next step is to outline the specific exercises...
    
    \vspace{0.5em}
    \hrule \vspace{0.5em}
    
    $\blacktriangleright$ \textbf{Memory Used:}\\
    \texttt{[Constraints: Study Sessions (Tue/Thu 20:00-21:30, Sat 10:00-12:00)]} $\rightarrow$ \texttt{[Validation]}: Check for potential time conflicts with newly proposed workout days.
\end{tcolorbox}

\subsection{Prompts}

\newtcolorbox{promptbox}[2][]{%
    colback=gray!5!white,      
    colframe=gray!75!black,    
    fonttitle=\bfseries,       
    title={#2},                
    sharp corners,             
    boxrule=0.8pt,             
    left=6pt, right=6pt, top=6pt, bottom=6pt, 
    #1
}

This section presents some of the important prompt templates involved in the paper.

\subsubsection{Evaluation Prompts}
Figures~\ref{fig:prompt_response_eval} and \ref{fig:prompt_memory_eval} illustrate the specific prompt templates used in our evaluation framework for assessing response consistency and memory retrieval quality, respectively.
\begin{figure*}[!t]
    \centering
    \begin{promptbox}{Prompt for Response Consistency Evaluation}
        \small
        \textbf{Instruction:} Your task is to evaluate the consistency between the \texttt{[candidate answer]} and the \texttt{[user-related memory]}.
        
        \vspace{0.5em}
        \textbf{Input Data:}
        \begin{itemize}[leftmargin=*, nosep]
            \item User’s current query
            \item User-related memory (representing the latest valid user state)
            \item Reference answer (based on the relevant memory)
            \item Candidate answer (to be evaluated)
        \end{itemize}

        \vspace{0.5em}
        \textbf{Evaluation Rules:}
        \begin{itemize}[leftmargin=*, nosep]
            \item Focus only on whether ``facts, constraints, preferences, and confirmed states'' are correctly used.
            \item Do NOT evaluate language style, tone, politeness, empathy, or fluency.
            \item Do NOT give a high score just because the answer ``sounds reasonable''.
            \item The reference answer is only to help understand how relevant memory should ideally be used; a candidate answer does not need to exactly match the reference answer to receive a full score.
        \end{itemize}

        \vspace{0.5em}
        \textbf{Scoring Criteria:}
        \begin{itemize}[leftmargin=*, nosep]
            \item \textbf{Score 0 (Poor):} The candidate answer conflicts with the user-related memory.
            \item \textbf{Score 1 (Fair):} The candidate answer does not conflict with the relevant memory but is generic and not based on user memory.
            \item \textbf{Score 2 (Good):} The candidate answer uses part of the user-related memory.
            \item \textbf{Score 3 (Very good):} The candidate answer (like the reference answer) uses all of the user-related memory.
        \end{itemize}

        \vspace{0.5em}
        \textbf{Output Format:}
\begin{verbatim}
{
    "score": int,
    "reason": str  # Briefly explain the reason for the score
}
\end{verbatim}
    \end{promptbox}
    \caption{The specific prompt used for evaluating the consistency between the generated response and user memory.}
    \label{fig:prompt_response_eval}
\end{figure*}

\begin{figure*}[!t]
    \centering
    \begin{promptbox}{Prompt for Memory Retrieval Quality Evaluation}
        \small
        \textbf{Instruction:} Your task is to evaluate the consistency between the \texttt{[retrieved memory]} and the \texttt{[ground-truth memory]}, and whether the retrieved memory is helpful.

        \vspace{0.5em}
        \textbf{Input Data:}
        \begin{itemize}[leftmargin=*, nosep]
            \item \texttt{<question>}: The user’s current query.
            \item \texttt{<groundtruth\_memory>}: The true memory that is helpful for answering the question.
            \item \texttt{<retrieved\_memory>}: The retrieved memory.
        \end{itemize}

        \vspace{0.5em}
        \textbf{Evaluation Dimensions:}
        \begin{enumerate}[leftmargin=*, nosep]
            \item \textbf{Memory Recall} (0--1): Semantics-aware memory recall calculation.
            \begin{itemize}[leftmargin=10pt, topsep=2pt]
                \item \textit{step1:} For each groundtruth\_memory, check in sequence whether its semantics are contained in any retrieved\_memory.
                \item \textit{step2:} Count how many groundtruth\_memory items are covered (hits\_cnt).
                \item \textit{step3:} Compute the final recall score as hits\_cnt / total number of groundtruth\_memory items.
            \end{itemize}
            
            \item \textbf{Memory Helpfulness} (0--2): The helpfulness of the retrieved memory.
            \begin{itemize}[leftmargin=10pt, topsep=2pt]
                \item \textbf{Score 0:} retrieved\_memory contains mutually conflicting or contradictory memories, which not only fail to help answer the question but may also cause confusion.
                \item \textbf{Score 1:} retrieved\_memory is somewhat helpful for answering the question (can provide partial supporting evidence).
                \item \textbf{Score 2:} retrieved\_memory is very helpful for answering the question (can provide comprehensive supporting evidence).
            \end{itemize}
        \end{enumerate}

        \vspace{0.5em}
        \textbf{Output Format:}
        Please provide your evaluation results using the following structure:
\begin{verbatim}
{
  "Mem_recall": float,
  "Mem_helpful_score": int,
  "Mem_hits": list[str],  # List the matched groundtruth_memory items
  "Mem_helpful_reason": str  # Explain the reason for the score
}
\end{verbatim}
    \end{promptbox}
    \caption{The specific prompt used for evaluating the quality of memory retrieval, focusing on semantic recall and helpfulness metrics.}
    \label{fig:prompt_memory_eval}
\end{figure*}

\end{document}